\documentclass[conference]{ieeeconf}
\IEEEoverridecommandlockouts
\usepackage{cite}
\usepackage{amsmath,amssymb,amsfonts}
\usepackage{algorithmic}
\usepackage{hyperref, graphicx}
\usepackage{textcomp}

\usepackage{multicol,lipsum,microtype}

\usepackage{xcolor}
\def\BibTeX{{\rm B\kern-.05em{\sc i\kern-.025em b}\kern-.08em
    T\kern-.1667em\lower.7ex\hbox{E}\kern-.125emX}}
    
\begin{document}

\title{\LARGE \bf Robotic Control of the Deformation of Soft Linear Objects \\ 
Using Deep Reinforcement Learning}

\author{Mélodie Hani Daniel Zakaria$^1$, Miguel Aranda$^2$, Laurent Lequièvre$^1$, Sébastien Lengagne$^1$, \\Juan Antonio Corrales Ramón$^3$ and Youcef Mezouar$^1$
\thanks{$^1$CNRS, Clermont Auvergne INP, Institut Pascal,
Université Clermont Auvergne, Clermont-Ferrand, France. $^2$Instituto de Investigación en Ingeniería de Aragón, Universidad de Zaragoza, Zaragoza, Spain. $^3$Centro Singular de Investigación en Tecnoloxías Intelixentes (CiTIUS),  Universidade de Santiago de Compostela, Santiago de Compostela, Spain. Corresponding author: Mélodie Hani Daniel Zakaria, e-mail: \texttt{Melodie.HANI\_DANIEL\_ZAKARIA@uca.fr.}}
\thanks{This work is funded by AURA through the ATTRIHUM project and from the EU Horizon 2020 research and innovation programme under grant agreement No 869855 (Project ’SoftManBot’). MA is funded via project PGC2018-098719-B-I00 (MCIU/AEI/FEDER, UE), and by the Spanish Ministry of Universities and the European Union-NextGenerationEU via a María Zambrano fellowship. JACR is funded by the Spanish Ministry of Universities through a ’Beatriz Galindo’ fellowship (Ref. BG20/00143) and by the Spanish Ministry of Science and Innovation through the research project PID2020-119367RB-I00.}
}

\maketitle

\begin{abstract}
This paper proposes a new control framework for manipulating soft objects. A Deep Reinforcement Learning (DRL) approach is used to make the shape of a deformable object reach a set of desired points by controlling a robotic arm which manipulates it. Our framework is more easily generalizable than existing ones: it can work directly with different initial and desired final shapes without need for relearning. We achieve this by using learning parallelization, i.e., executing multiple agents in parallel on various environment instances. We focus our study on deformable linear objects. These objects are interesting in industrial and agricultural domains, yet their manipulation with robots, especially in 3D workspaces, remains challenging. We simulate the entire environment, i.e., the soft object and the robot, for the training and the testing using PyBullet and OpenAI Gym. We use a combination of state-of-the-art DRL techniques, the main ingredient being a training approach for the learning agent (i.e., the robot) based on Deep Deterministic Policy Gradient (DDPG). Our simulation results support the usefulness and enhanced generality of the proposed approach.
\end{abstract}


\section{Introduction}
The manipulation of deformable objects is currently a relevant topic in robotics research \cite{Yin2021SR, Sanchez2018Survey}. In particular, the manipulation of Deformable Linear Objects (DLOs) has high relevance in automation applications: examples of interesting tasks that have been addressed include cable harnessing \cite{Zhu2018,Lagneau2020}, USB wire soldering \cite{Gao2020}, or vegetable plant manipulation \cite{Botterill2017}. One possible perspective on this problem is to study model-based manipulation planning, as done in \cite{Bretl2014,Mukadam2014} for elastic rods. In this paper, we are instead interested in the online control of the robot to deform a DLO in a desired way in conditions of high uncertainty and with no knowledge of the object's mechanical deformation model. The works that addressed a similar scenario considered mostly 2D workspaces \cite{Zhu2018,Qi2021,Koessler2021}, while control in 3D is significantly more challenging due to the higher complexity of object modeling and perception. Some works addressed control in 3D for small deformations \cite{Navarro-Alarcon2016TR,Lagneau2020}. Overall, while \textit{classical} methods have achieved important progress in this field, the existing challenges motivate us to explore a solution based on Deep Reinforcement Learning (DRL) \cite{Laezza2021ICRA}.

\begin{figure}[ht]
    \centering
    \includegraphics[width=8cm]{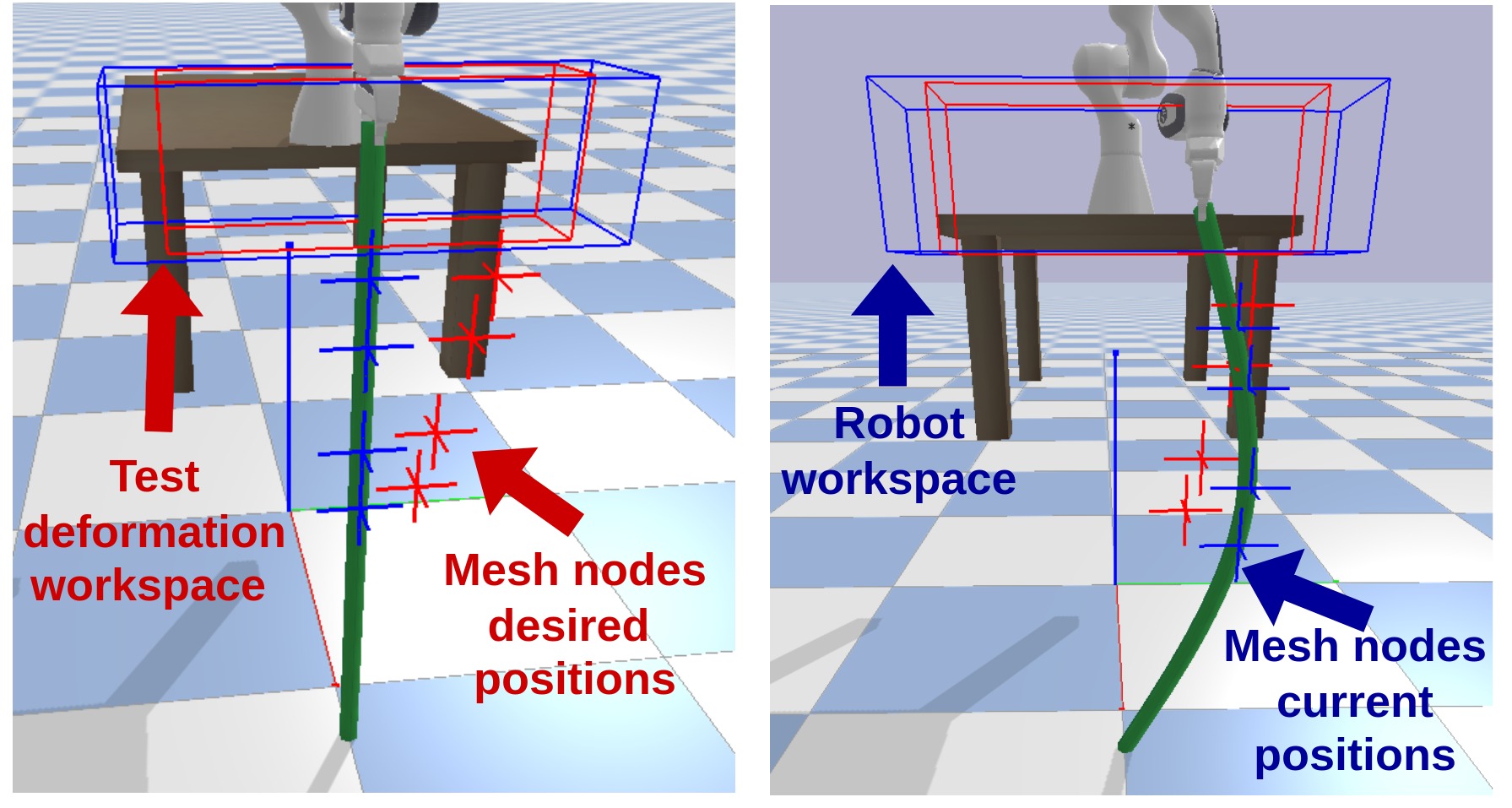}
    \vspace{-2mm}
    \caption{The setup we consider, including an illustration of some elements of our solution. The robot deforms the soft linear object (green) by making the selected mesh nodes (i.e., the blue points) reach the desired corresponding positions (i.e., the red points). The points are marked as crosses. The robot tip position has to remain within the deformation workspace (i.e., the red box) for performing the desired deformation. The deformation workspace used in testing is bigger than the training workspace. The blue box delimits the robot's workspace, i.e., the robot's gripper tip cannot reach a position out of that box due to the robot's articular limits.}
    \label{Deformable_object_simulation}
    \vspace{-2mm}
\end{figure}

The robotics community has increasingly adopted the usage of DRL algorithms to control robots\cite{Zhao2020SSCI}. Most of these works involve working with rigid bodies with no or negligible deformations\cite{Matas2018CRL, Yin2021SR}. However, soft object manipulation has many crucial applications, especially in household robotic assistance, medicine, and industry\cite{Jangir2020ICRA, Matas2018CRL}. In industrial automation, DRL has already been identified as interesting in tasks with high modeling uncertainty and need for high dexterity. For instance \cite{Hou2020,Leyendecker2021} used reinforcement learning for industrial assembly, albeit without having to deal with deformable objects as we do here.

In the literature, the works based on DRL for manipulating deformable objects are, on the one hand, only formulated for simple tasks\cite{Yin2021SR} such as hanging a cloth\cite{Jangir2020ICRA, Matas2018CRL} or moving a rope\cite{Laezza2021ICRA}. On the other hand, most of the soft objects used are 2D\cite{Yin2021SR}: the mesh used to model the object is a 2D mesh, i.e., formed by 2D polygons such as triangles. Promoting progress in this regard, SoftGym\cite{Lin2020softgym} presented a set of benchmarks for manipulating soft objects (including 3D objects) using OpenAI Gym\cite{Brockman2016OpenAI} and Python interface.

The main drawback of the existing techniques, whether used in simulation\cite{Laezza2021ICRA, Jangir2020ICRA} or in real experiments\cite{Matas2018CRL}, is that they are not easily generalizable\cite{Laezza2021ICRA,Yin2021SR,Zhao2020SSCI}. Their agent is trained to perform a manipulation from constant initial to constant target deformations, and it is not trained to deal with different configurations. As an example in DLO manipulation, in \cite{Laezza2021ICRA} the authors control the object shape from some initial states to some desired deformations that are not changeable.


This paper describes a new framework for the robotic control of the shape of DLOs. We use a combination of state-of-the-art DRL algorithms and techniques to build up our framework. We use learning parallelization to make our framework generalizable, i.e., we execute multiple agents in parallel on various environment instances. We focus our study on deformable linear objects. The contributions of our framework are:
\begin{enumerate}
    \item Its generalizability, i.e., we train the agent only once (using a specific soft object), and it can deform the soft object starting from a different initial position and end up with a different desired shape. Moreover, the agent can make the soft object reach an untrained position, i.e., we train the agent on a small workspace and test it on a bigger one.
    \item The complexity of the accomplished task. As shown in Figure~\ref{Deformable_object_simulation}, the robot deforms a foam bar by making some selected mesh nodes reach the correspondent desired positions in 3D space, potentially involving complex torsion motions. This is made possible by modeling the object with a 3D tetrahedral mesh and via our DRL system design.
    
\end{enumerate}
We train and evaluate our approach in simulation. Our evaluation is carried out in diverse conditions and it validates the capability of the proposed approach.

\section{Problem statement}\label{sec_problem_statement}
We address the problem of controlling the deformation of a DLO using a robot arm that manipulates it. For simplicity, the robot grasps one end of the object, and the other end is fixed to the ground. The object is represented by a mesh and we describe its deformation by a set of selected mesh nodes. The objective is to control the arm so that the positions of the selected nodes are driven to prescribed values. The difficulty of this indirect control problem lies in the fact that the dynamical model of the system to be controlled is complex and uncertain. We propose a generalizable architecture to solve this problem based on DRL. The problem setup is illustrated in Figure \ref{Deformable_object_simulation} and our solution will be detailed in the remainder of the paper. 

\section{Background on Soft Object Manipulation using DRL} \label{Related Work}
This section gives background on the problem of soft object manipulation using DRL. We will focus on discussing aspects that are particularly important for our application.


\subsection{Representing deformable object shape}
The most widely used technique is to represent the soft object shape through images\cite{Jangir2020ICRA} instead of modeling it since it is challenging to have a precise model\cite{Yin2021SR}. In \cite{Matas2018CRL}, a Neural Network detects the soft object shape thanks to supervised learning. The disadvantage of using images is that the computational cost increases, and it is hard to learn afterward (i.e., via DRL) because the resulting state-space is large\cite{Jangir2020ICRA}. In \cite{Laezza2021ICRA}, a method based on geometry calculations is proposed to represent the object shape. Another method is based on selecting some mesh nodes in the object model describing the deformation and using their positions as state-space inputs\cite{Yin2021SR, Jangir2020ICRA}. We preferred to use the latter technique because it is easier to set up, and it keeps the size of the space-state relatively small, which facilitates the training.


\subsection{Techniques to deal with the manipulation complexity}
The most common technique in the state-of-the-art is to combine imitation learning with reinforcement learning\cite{Yin2021SR}. Imitation learning is used to reduce the complexity of the manipulation by using demonstrations given by an expert. Another method that we mentioned previously is to have a detailed perception of the object's shape through images\cite{Jangir2020ICRA}. The drawback of both methods is that they have a high computational cost and a large state-space\cite{Laezza2021ICRA, Jangir2020ICRA}. We prefer to use only a DRL algorithm and select a few mesh nodes that describe the deformation of the object as input to the state-space. This way, our state-space is small, which makes learning easier.

\subsection{Physics-based simulator}
Usually, the training of the agent is done in simulation, using a physics-based simulation engine \cite{Zhao2020SSCI}. OpenAI Gym\cite{Brockman2016OpenAI} defines an architecture with the main components needed to train the agent, such as resetting the environment, making an action, getting an observation of the state of the environment, and computing the reward. The environment created on the simulator has to have such components. The most popular simulators for deformable object manipulation in the robotics community are MuJoCo\cite{Todorov2012MuJoCo} and Bullet\cite{Coumans2021Pybullet}. We prefer to use PyBullet, the Python interface of Bullet, because it is powerful and open-source.



\section{Components of our DRL framework} \label{Background}
In this section, we present standard components of DRL systems that we use in our framework. We focus on explaining how we incorporate them in the framework. The elements include the RL procedure, the Bellman equation, the DDPG algorithm, and the reward function. 

\begin{figure*}[ht]
\centering
    \includegraphics[width=\textwidth]{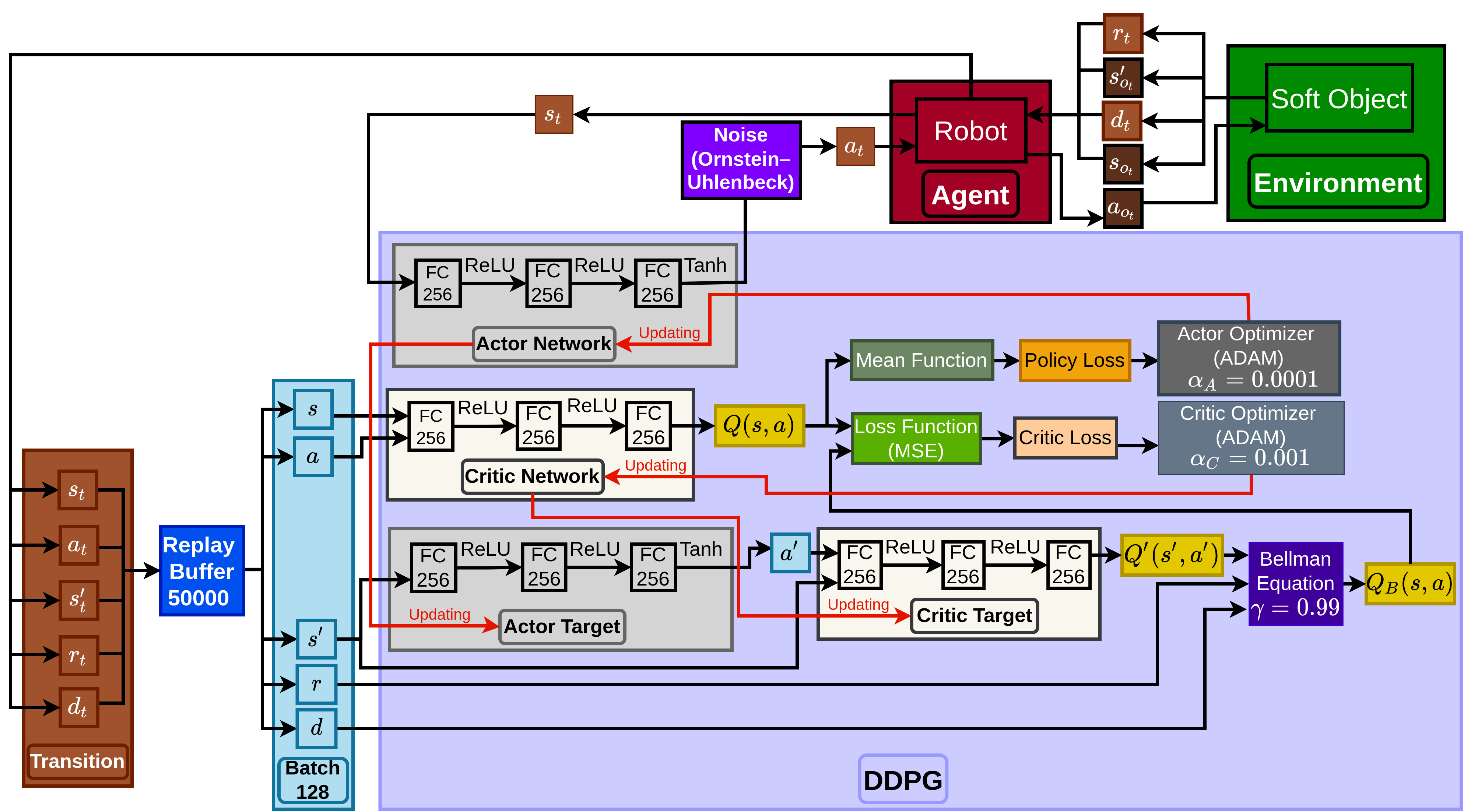}
\caption{Overview of our proposed framework for controlling the deformation of a soft object via the DDPG algorithm. The structure of the full DRL system and relevant parameters are displayed.}
\label{Rl_algo}
\vspace{-5mm}
\end{figure*}

\subsection{RL procedure}
We consider a classical trial-and-error RL procedure consisting of an agent (e.g., robot) interacting with the environment (e.g., the soft object) based on the policy to maximize rewards on discrete timesteps\cite{Lillicrap2015DDPG}. In each transition $t$, the agent starts from the state $s_t$, and takes an action $a_t$, which changes the state to a next state $s^\prime_t$\cite{Zakaria2021Frontiers}. The state $s_t$ and the action $a_t$ are included in the continuous state space $\mathcal{S}$, and the continuous action space $\mathcal{A}$, respectively, i.e., $s_t \in \mathcal{S}$ and $a_t \in \mathcal{A}$.

The observation the agent got from the environment describes the changes that happened by moving from state $s_t$ to $s^\prime_t$. The reward $r_t$ evaluates the action taken $a_t$ according to the task goal. The agent's goal is to learn the optimal policy $\pi^*: \mathcal{S} \longrightarrow \mathcal{A}$ throughout the different transitions. A transition $t$ is made of an action $a_t$, a state $s_t$, a next state $s_t'$, a reward $r_t$, and a variable called done, $d_t$, that expresses if the action achieved the task goal $(d_t = 1)$ or not $(d_t = 0)$.

\subsection{Bellman equation}
The Bellman equation\cite{Nachum2017ANIPS} is used to calculate a Q-value $Q_{B_t}(s_t,a_t)$ that evaluates the action $a_t$ chosen in a current state $s_t$. The Bellman equation (\ref{bellman_equation}) considers the discount factor $(\gamma \in [0.9,1])$ and the next Q-values $Q^\prime_t(s^\prime_t,a^\prime_t)$ to calculate $Q_{B_t}(s_t,a_t)$. The discount factor controls how much the DRL learning is considering future rewards. The next Q-values  $Q^\prime_t(s^\prime_t,a^\prime_t)$ are calculated for choosing the next action $a^\prime_t$ in the next state $s^\prime_t$.
\begin{equation} \label{bellman_equation}
   Q_{B_t}(s_t,a_t) = r_t + \gamma \times Q^\prime_t(s^\prime_t,a^\prime_t) \times (1-d_t).
\end{equation}


\subsection{Off-policy vs on-policy learning}
We choose to use an off-policy (as opposed to on-policy) algorithm because it allows parallelizing the learning\cite{Hausknecht2016AAAI}. Parallelizing the learning means executing multiple agents in parallel on various environment instances. The learning parallelization technique speeds up the convergence, i.e., learning can be faster\cite{Kretchmar2002CSCI, Clemente2017arXiv}. We will discuss this concept further in Section~\ref{parallelization}.

\subsection{Deep Deterministic Policy Gradient (DDPG)}
The DDPG is a DRL algorithm based on Actor-Critic methods used for dealing with continuous action spaces\cite{Lillicrap2015DDPG}. It learns a Q-function and a policy by utilizing off-policy data and the Bellman equation\cite{Jangir2020ICRA}. The actor network (policy network) has as input the state $s_t$ and gives as output the optimal action $a_t$. The critic network (Q-function network) evaluates the optimality of the action $a_t$ chosen at state $s_t$ by attributing it the Q-values $Q_t(s_t,a_t)$ at transition $t$. Figure~\ref{Rl_algo} presents an overview of the framework established to make the robot deform a soft object using the DDPG algorithm. Next, we detail the modules in the algorithm.

\subsubsection{Pre-training procedure}
The agent applies the action $a_t$ selected by the actor network within the state $s_t$ to the environment in order to store the inputs of the environment ($a_t$ selected in $s_t$) and its outputs ($s^\prime_t$, $r_t$, and $d_t$) that constitute the transition $t$ in the replay buffer (cf. Figure~\ref{Rl_algo}). The training of the actor and critic networks can only begin once the replay buffer contains enough transitions to extract a batch. A batch is composed of elements (i.e., actions $a$, states $s$, next states $s^\prime$, etc.) coming from several non-sequential transitions. These transitions are selected randomly.

\subsubsection{Training procedure}
Making the agent learn from previous memories, i.e., using batches, accelerates learning and breaks undesirable temporal correlations\cite{Liu2018AACCCC}. The training of the critic network consists of reducing the error between the Q-values calculated using the Bellman equation $Q_B(s,a)$ (cf. (\ref{bellman_equation})) and the Q-values estimated by the critic network $Q(s,a)$ (cf. Figure~\ref{Rl_algo}). The Q-values number equals the batch size $N$, i.e., the number of selected transitions to train the agent. The Mean Square Error (MSE) optimization technique is used to reduce that error, i.e., we use the following Critic loss:
\begin{equation}
    \text{Critic loss} = \text{MSE}(Q_B(s,a),Q(s,a)).
\end{equation}

The weights of the critic network are updated based on the Critic loss. The ADAM optimizer\cite{Kingma2014ADAM} is used to calculate the gradient descent. The Q-values given by the critic network $Q(s,a)$ are used to evaluate the actions $a$ chosen by the actor at states $s$. Then, the actor's training is based on the Q-value given by the critic network, i.e., the actor loss is equal to the Q-value. Since the agent's training is made from a batch, one obtains as many Q-values $Q(s,a)$ (cf. Figure~\ref{Rl_algo}) as there are transitions in the batch. The policy loss is calculated by taking an average of the $Q(s,a)$ \cite{Lillicrap2015DDPG}:
\begin{equation}\label{MSE}
    \text{Policy loss} = -\overline{Q(s,a)} = - \frac{\sum_{t=1}^{N} Q_t(s_t,a_t)}{N}.
\end{equation}
The weights of the actor network are updated based on the policy loss.

\subsubsection{Target networks}
Using a target network is a technique to stabilize learning. A target network is a copy of the main network's weights held constant to act as a stable target for learning for a fixed number of timesteps\cite{Lillicrap2015DDPG}. We use Polyak averaging to update the target networks (also called soft updating) once per the main network's update\cite{Sehgal2019ICR}:
\vspace{-1mm}
\begin{equation}
    W_{A_T} = \tau W_A + (1-\tau) W_{A_T}
\end{equation}
\vspace{-6mm}
\begin{equation}
    W_{C_T} = \tau W_C + (1-\tau) W_{C_T},
\end{equation}
where the used terms are:
\begin{itemize}
    \item $W_{A_T}$: the weights of the actor target network.
    \item $W_A$: the weights of the actor network.
    \item $W_{C_T}$: the weights of the critic target network.
    \item $W_C$: the weights of the critic network.
    \item $\tau$: the Polyak factor.
\end{itemize}

We choose to utilize the DDPG algorithm as our DRL algorithm because it is suitable for continuous action spaces. It has fewer parameters to set than other actor-critic DRL algorithms. It is a powerful tool to generalize the training, combined with parallel learning.

\subsection{Reward function}
The reward function is the key element that allows us to control and optimize the agent policy of choosing actions\cite{Zakaria2021Frontiers}. More details about choosing the suitable reward function are given in \cite{Sutton2018MIT, Plappert2018arXiv}. The simplest dense reward function for our task is to use a Euclidean distance-based calculation\cite{Laezza2021ICRA}. Therefore, our reward $r_t$ is calculated as the average Euclidean distance between the current positions of the selected mesh nodes, and their desired positions.


\subsection{Learning parallelization} \label{parallelization}
The actor-critic DRL algorithm A3C\cite{Mnih2016A3C} proposes to asynchronously execute multiple agents in parallel on various instances of the environment. That parallelism decorrelates the agent learning data since, at any transition, the parallel agents will be experiencing a variety of different states. Combining batch extraction and learning parallelization for off-policy algorithms ensures that the training data are decorrelated and can be collected faster\cite{Kretchmar2002CSCI, Clemente2017arXiv}. Thus, combining both techniques improves the overall learning time while achieving a better result from the generalization point of view. That is why we train the agent using DPPG on a single multi-core CPU, as in \cite{Mnih2016A3C}.


\section{Framework Implementation} \label{Methods}
After having introduced all the necessary DRL components, we describe how we apply them in our novel framework to address the specific problem scenario considered (Section \ref{sec_problem_statement}). We provide the implementation details including all assigned values for the DDPG and simulation parameters.


\subsection{Framework overview}
Before starting the learning phase, we create a deformation space box within which the robot gripper tip moves to deform the object, and we record the positions of the selected nodes $P_d$ in a database. The reason for using a deformation space box is to record several deformations within a limited space that is reachable by the gripper tip. We have created two databases, each based on a box of different size: the training one, which is smaller, and the testing one, which is larger. All the details about the databases are mentioned in Section~\ref{sim_params}. The robot's objective is to manipulate the object so that the current positions of the selected nodes $P_c$ reach the desired positions $P_d$ within a tolerance threshold.

Figure~\ref{Rl_algo} gives an overview of our architecture. The action $a_t$ given by the DDPG to the agent (i.e., the robot) is the Cartesian velocity of the gripper tip $a_t \in \mathcal{A} = (V_x, V_y, V_z) \implies a_t \in \mathbb{R}^3$. The action $a_t$ is continuous since each element of the velocity ($V_x, V_y, \text{ or } V_z$) can have any value within the interval $ [-1,1]$. Then, the action $a_t$ is integrated according to the timestep (which is equal to \mbox{0.06 s}) to calculate the new gripper tip position $(X_n, Y_n, Z_n)$. The classical position-based controller available in Bullet moves the arm from its current position $(X_c,Y_c,Z_c)$ to the new one $(X_n,Y_n,Z_n)$.

The state $s_t \in \mathcal{S}$ is made up of the gripper tip current state $s_{g_t} \in \mathbb{R}^6$ and the current object shape $s_{o_t} \in \mathbb{R}^{6m}$ (cf. (\ref{state})) with $m$ the number of selected mesh nodes. $s_{g_t}$ includes the gripper tip position $(X_c,Y_c,Z_c)$ and velocity $(V_x, V_y, V_z)$. $s_{o_t}$ is composed of the positions of the selected mesh nodes $P_c \in \mathbb{R}^{3m}$, and their desired positions $P_d \in \mathbb{R}^{3m}$.
\begin{equation}\label{state}
    s_t = (s_{g_t}, s_{o_t}) \in \mathcal{S} = (X_c,Y_c,Z_c, V_x, V_y, V_z, P_c,  P_d).
\end{equation}
We calculate the reward $r_t$ as the average Euclidean distance $D_t$ between the current positions of the selected mesh nodes $P_c$ and their desired positions $P_d$ (cf. (\ref{ShapedReward})). Using subindex $i$ to denote the position of a single mesh node, we have:
\begin{equation} \label{ShapedReward} 
    r_t = -\overline{D_t(P_c,P_d)} = - \frac{\sum_{i=1}^{m} D_t(P_{c_i},P_{d_i})}{m}.
\end{equation}

\subsection{DDPG parameters}
The actor, actor target, critic, and critic target Deep Neural Networks (DNNs) have the same architecture: 3 Fully Connected (FC) hidden layers, each of which comprises 256 neurons. We use the Rectified Linear Unit (ReLU) as an activation function. We apply the Tanh function on the actor outputs $a_t$ to ensure that the gripper tip velocities remain in the interval $[-1,1]$. We add noise to the action $a_t$ using Ornstein–Uhlenbeck noise\cite{Lillicrap2015DDPG} for the exploration. We initialize the DNNs of the actor and critic with random values as in \cite{Lillicrap2015DDPG}. The actor target and critic target DNNs weights copy those of the actor and critic DNNs. The ADAM optimizer is used for gradient updates with learning rates of $\alpha_A = 0.0001$ for the actor and $\alpha_C = 0.001$ for the critic. A batch of 128 transitions is randomly sampled from the replay buffer, containing 50000 transitions. We use a constant discount factor $\gamma = 0.99$ and a constant Polyak factor $\tau = 0.01$.

Since we use parallel learning, in each episode, 32 agents are trying to achieve a different deformation during 300 transitions. This means that each agent makes 300 actions and passes through 300 different transitions to try to achieve 32 different goals (each agent has a different goal). Each action will have a reward $r_t$, and each agent will have a global reward equal to the sum of the action reward over the 300 transitions. This leads to having different gradients that are synchronized among the 32 agents, i.e., there will be one final gradient equal to the sum of all the 32 gradients. Then 32 agents networks are updated based on that final gradient so that all these networks keep having the same updated weights. We train the 32 agents during 63 episodes, which are equivalent to $32*63 = 2016$ episodes if we use a single agent and do not parallelize the training. The training lasts from 1000 to 1 million episodes in the literature\cite{Jangir2020ICRA, Laezza2021ICRA}. For training 32 agents, we used 32 CPU cores and the Python library MPI\cite{Dalcin2021MPI}. All the conducted training lasted less than three and a half hours.

\subsection{Simulation parameters}\label{sim_params}
We use PyBullet as physics engine of the simulator to train our agent. The simulator's physics engine uses the FEM method to simulate the soft object. The model of our soft object is built up from a 3D tetrahedral mesh. That model comprises 200 nodes, 392 tetrahedrons, 789 links, and 396 faces. The soft object has the following mechanical parameters: the Young's modulus is equal to 2.5 MPa, the Poisson coefficient is equal to 0.3, the mass is equal to 0.2 Kg, the damping ratio is equal to 0.01, and the friction coefficient is equal to 0.5. The simulation timestep is equal to 0.003 s. Note that we chose the timestep to integrate the action $a_t$ as equal to $20*0.003=0.06$ s., i.e., sufficiently larger than the simulation timestep.

\begin{figure}[ht]
    \centering
    \includegraphics[width=8cm]{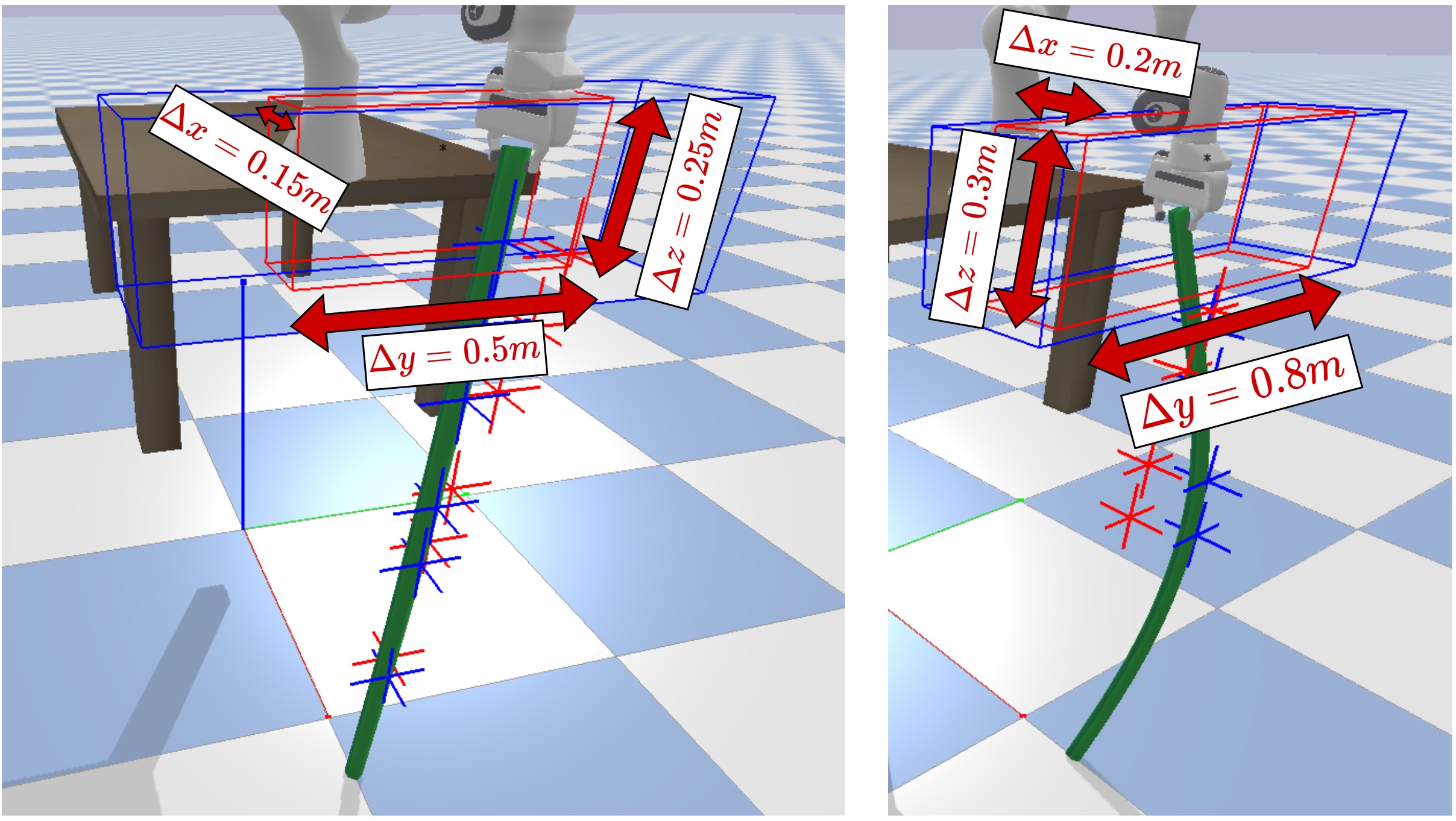}
    \caption{The left plot presents the small box (in red) used to generate the small database. The right plot shows the large box (in red) used to generate the large database.}
    \label{Deformation_box}
    \vspace{-6mm}
\end{figure}

We created two databases, each based on a box of different 3D size. The gripper tip moves inside that box to deform the object, and we recorded those deformations to use them as desired positions $P_d$ in the training and the testing phase. Figure~\ref{Deformation_box} shows the small and the large boxes. The small box size is equal to: 0.15 m on the x-axis, 0.5 m on the y-axis, and 0.25 m on the z-axis. The large box size is equal to: 0.2 m on the x-axis, 0.8 m on the y-axis, and 0.3 m on the z-axis. The small box is used to generate the small database, which contains 930 deformations. The large box is used to generate the large database, which includes 2651 deformations. The small database is used for the training, and both databases are used for the testing.


\section{Experimentation} \label{Experimentation}
\begin{figure}[ht]
    \centering
    \includegraphics[width=8cm]{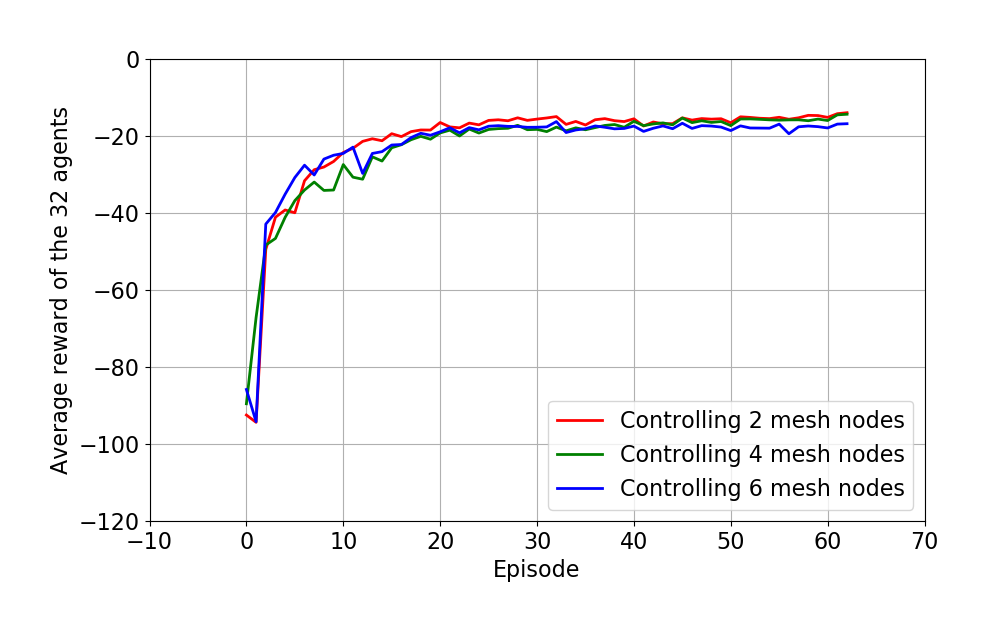}
    \vspace{-3mm}
    \caption{Average reward obtained by the 32 agents in each episode when controlling two mesh nodes, four mesh nodes, and six mesh nodes.}
    \label{learning_curve}
    \vspace{-2mm}
\end{figure}

This section presents our experimental results. We have done three trainings, to control: two mesh nodes, four mesh nodes, and six mesh nodes. We used an average distance error threshold of 0.05 m. As mentioned in the previous section, the training was parallelized: 32 agents were trained each for 63 episodes, leading to having 2016 episodes in total. We trained the agent using the small database to extract the desired deformations, i.e., the desired positions $P_d$ of the mesh nodes. During the training, the environment was reset to the initial configuration (the robot and the object returned to their initial position) after each episode. Figure~\ref{learning_curve} shows the average reward obtained by the 32 agents in each episode when controlling two mesh nodes, four mesh nodes, and six mesh nodes. As we can notice from Figure~\ref{learning_curve}, there is no need to smooth the learning curves, as in the literature\cite{Jangir2020ICRA, Laezza2021ICRA}. This is thanks to the stability of the learning due to its parallelization.

\vspace{-4 mm}

\begin{figure*}[ht]
\centering
    \includegraphics[width=\textwidth]{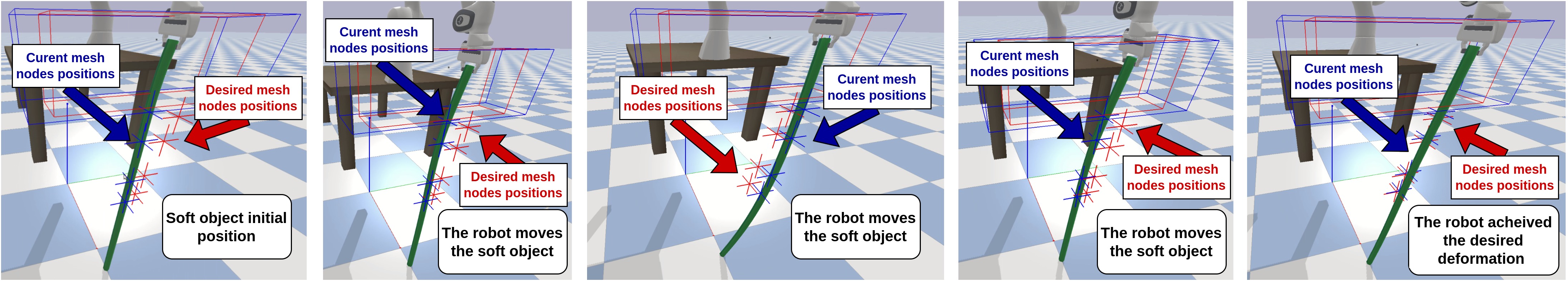}
    \vspace{-7mm}
\caption{Example of the robot deforming the soft object: four mesh nodes reach their desired positions with an average distance error threshold of 0.05 m.}
\label{deformation example}
\vspace{-1mm}
\end{figure*}

\begin{table*}[ht]
\centering
\scalebox{0.8}{{\renewcommand{\arraystretch}{1.5} 
\begin{tabular}{|p{2.7cm}|p{2.3cm}|p{1cm}|p{3.2cm}|p{1.2cm}|p{4.2cm}|p{1cm}|}
\hline
 & Mesh nodes number & Database & Testing error threshold (m) & Done (\%) & Mean average distance error $\pm \sigma$ (m)  & Best (m) \\
\hline
\centering With reinitialization & \centering 2 & \centering Small & \centering 0.05 & \centering 99.5 & \centering 0.03010 $\pm$ 0.00591 & 0.01156 \\
    \cline{4-7}
    & & & \centering 0.03 & \centering80.2 & \centering 0.02987 $\pm$ 0.00993 & 0.01156\\
        \cline{3-7}
        & & \centering Large & \centering 0.05 & \centering 73.2 & \centering 0.05886 $\pm$ 0.02730 & 0.02724\\
            \cline{4-7}
             & & & \centering 0.03 & \centering 21.1 & \centering 0.06866 $\pm$ 0.03214  & 0.01862 \\
    \cline{2-7}
    & \centering 4 & \centering Small & \centering 0.05 & \centering 99.8  & \centering 0.04251 $\pm$  0.00631 & 0.01012 \\
        \cline{4-7}
        & & & \centering 0.03 & \centering 88.7 & \centering 0.02816 $\pm$ 0.00758 & 0.01012 \\
            \cline{3-7}
            & & \centering Large & \centering 0.05 & \centering 97.2 & \centering 0.04386 $\pm$ 0.00840 & 0.02130\\
            \cline{4-7}
            & & & \centering 0.03 & \centering 46.8 & \centering 0.04650 $\pm$ 0.02306  & 0.01309 \\
    \cline{2-7}
    & \centering 6 & \centering Small & \centering 0.05 & \centering 99.1 & \centering 0.04473 $\pm$ 0.00685 & 0.01009 \\
        \cline{4-7}
             & & & \centering 0.03 & \centering 73.4 & \centering 0.02796 $\pm$ 0.01183 & 0.01009  \\
            \cline{3-7}
            & & \centering Large & \centering 0.05 & \centering 79.0 &  \centering 0.05144 $\pm$ 0.01506 & 0.02779 \\
                \cline{4-7}
                & & & \centering 0.03 & \centering 20.0 & \centering 0.06415 $\pm$ 0.02563 & 0.01548 \\
\hline
\centering Without reinitialization & \centering 2 & \centering Small & \centering 0.05 & \centering 87.9 & \centering 0.04530 $\pm$  0.01142 & 0.01579  \\
    \cline{4-7}
    & & & \centering 0.03 & \centering 47.5 & \centering 0.04107 $\pm$ 0.01808 & 0.01236 \\
        \cline{3-7}
            & & \centering Large & \centering 0.05 & \centering 44.9 & \centering 0.07411 $\pm$ 0.04454 & 0.01642 \\
                \cline{4-7}
                & & & \centering 0.03 & \centering 13.9 & \centering 0.08038 $\pm$ 0.04536 & 0.02438 \\
    \cline{2-7}
    & \centering 4 & \centering Small & \centering 0.05 & \centering 93.7 & \centering 0.04486 $\pm$ 0.01215  & 0.00569  \\
        \cline{4-7}
        & & & \centering 0.03 & \centering 45.2 & \centering 0.05015 $\pm$ 0.02732 & 0.01399 \\
            \cline{3-7}
            & & \centering Large & \centering 0.05 & \centering 72.8 & \centering 0.05365 $\pm$ 0.02201 & 0.01506 \\
            \cline{4-7}
            & & & \centering 0.03 & \centering 21.0 & \centering 0.05873 $\pm$ 0.02551  & 0.01343 \\
    \cline{2-7}
    & \centering 6 & \centering Small & \centering 0.05 & \centering 36.8 & \centering 0.08669 $\pm$ 0.03817 & 0.02106 \\
        \cline{4-7}
        & & & \centering 0.03 & \centering 15.7 & \centering 0.07478 $\pm$ 0.03371  & 0.01826 \\
            \cline{3-7}
            & & \centering Large & \centering 0.05 & \centering 64.8 & \centering 0.05914 $\pm$ 0.02447  & 0.02771 \\
                \cline{4-7}
                & & & \centering 0.03 & \centering 15.2 & \centering 0.06244 $\pm$ 0.02655 & 0.01537 \\
\hline
\end{tabular}}}
\caption{Results of all the conducted tests}
\vspace{-8.5mm}
\label{results_table}
\end{table*}

For the testing phase, all the results are calculated for 1000 testing episodes with 30 steps, i.e., the robot can take a maximum of 30 actions to achieve the deformation. We test the three trainings for an average distance error threshold of 0.05 m and 0.03 m. We evaluate them using the small and the large databases to extract the desired deformations. We assess them finally with and without reinitializing the environment. All these results are presented in Table~\ref{results_table}. In Table~\ref{results_table}, the column "done" indicates the percentage of the agent's success to achieve the desired deformations. The percentage is calculated on the 1000 episodes with 30 steps. The "Best" column reveals the minimum distance error obtained within the 1000 episodes.

Figure~\ref{deformation example} presents an example of the robot deforming a soft object to reach a new deformation on which the robot was not trained. Other deformations are presented in the video available on \url{https://youtu.be/MbFCS59ZZ_4}. As we can notice from Table~\ref{results_table}, in the case that we reinitialize the environment and use the same database and distance error threshold as in training, the agent achieves in the worst-case scenario \mbox{99.1 \%} deformations. If we only change the distance error threshold to 0.03 m, the agent succeeds in attaining in the worst-case scenario \mbox{73.4 \%} deformations. If we only change the database to the large one, the agent realizes in the worst-case scenario \mbox{73.2 \%} deformations. For the last test, we do not reinitialize the environment and we keep the other parameters constant. Specifically, the initial position of the soft object in the current episode is the desired one achieved by the robot in the previous episode. Therefore, in order to succeed in this scenario the agent needs to have learnt a stronger, more general policy. In this more challenging scenario, the robot succeeds in making \mbox{87.9 \%} deformations while controlling two mesh nodes, \mbox{93.7 \%} deformations while controlling four mesh nodes, and \mbox{36.8 \%} deformations while controlling six mesh nodes. We can observe that the results for four mesh nodes are better than for two mesh nodes. Our interpretation is that describing the deformation of an object using only two mesh nodes is not precise enough, hence the agent has difficulty generalizing what it has learned during training.

The results prove that our framework is generalizable. We trained the agent using a small deformations database, with a constant distance error threshold and reinitializing the environment after each episode. The agent can be more precise in the testing phase than in training, as shown by our tests with lower distance error threshold. The agent achieves other deformations than those used during training, without needing to be retrained. The agent makes the soft object reach the desired deformation even if the object position is not reinitialized. Our method presents a limitation when we combine the changes in the testing phase. Sometimes it performs well, such as when we test the four mesh nodes control on the large database without reinitializing the environment: in this case the robot achieves \mbox{72.8 \%} deformations. Sometimes the test fails, such as when we test the two mesh nodes control on the large database with a distance error threshold of 0.03. The robot achieves only \mbox{21.1 \%} deformations in this case. 

The entire code and results of tests conducted on another database are available on \url{https://github.com/MelodieDANIEL/robotic\_control\_of\_DLO\_using\_DRL}.



\section{Conclusion and future work} \label{Conclusion}
We have proposed a new control framework for the manipulation of soft objects, addressing a DLO deformation control task. We have assessed through experiments that our framework is generalizable, i.e., the agent can deform the soft object starting from a different initial position and end up with a different desired shape without having to relearn. We verified this by training the agent on a small deformations database and testing it on a large deformations database.


We note that there are still some points to improve in future work. Firstly, we want to use transfer learning techniques for sim-to-real, such as domain randomization, to test our framework on real experiments. Secondly, we want to evaluate our framework on other types of deformable objects.


\bibliographystyle{ieeetr}
\bibliography{bibliography}

\end{document}